\documentclass[main, biber]{now-journal}
\usepackage{epstopdf}
\usepackage{multirow}
\usepackage{makecell}
\bibliography{main}

\title{DefakeHop++: An Enhanced Lightweight Deepfake Detector}

\author[1]{Hong-Shuo Chen}
\author[2]{Shuowen Hu}
\author[2]{Suya You}
\author[1]{C.-C. Jay Kuo}

\affil{University of Southern California, Los Angeles, California, USA}

\affil{Army Research Laboratory, Adelphi, Maryland, USA}

\articledatabox{ISSN 2161-1823; DOI 10.1561/103.00000003\\
\copyright \ Hong-Shuo Chen}

\keywords{Deepfake detection, green learning, green AI, 
weakly-supervised learning.}

\begin{document}

\begin{abstract}

On the basis of DefakeHop, an enhanced lightweight Deepfake detector
called DefakeHop++ is proposed in this work.  The improvements lie in
two areas.  First, DefakeHop examines three facial regions (i.e., two
eyes and mouth) while DefakeHop++ includes eight more landmarks for
broader coverage. Second, for discriminant features selection, DefakeHop
uses an unsupervised approach while DefakeHop++ adopts a more effective
approach with supervision, called the Discriminant Feature Test (DFT).
In DefakeHop++, rich spatial and spectral features are first derived
from facial regions and landmarks automatically. Then, DFT is used to
select a subset of discriminant features for classifier training. As
compared with MobileNet v3 (a lightweight CNN model of 1.5M parameters
targeting at mobile applications), DefakeHop++ has a model of 238K
parameters, which is 16\% of MobileNet v3.  Furthermore, DefakeHop++
outperforms MobileNet v3 in Deepfake image detection performance in a
weakly-supervised setting. 

\end{abstract}

\section{Introduction}\label{sec:introduction}

It is common to see fake videos appearing in social media platforms
nowadays due to the popularity of Deepfake techniques.  Several mobile
apps can help people create fake contents without any special editing
skill.  Generally speaking, deepfake programs can change the identity of
one person in real videos to another realistically and easily.  The
number of Deepfake videos surges rapidly in recent years.  Fake videos
may result in serious damages to our society since people can be fooled
by them delivered by the Internet and some misinformation may make the
public panic and anxious. To address this emerging threat, it is
essential to develop lightweight Deepfake detectors that can be deployed
in mobile phones. 

With the fast growing Generative Adversarial Network (GAN) technology,
image forgery techniques keep evolving in recent years. They are
effective in reducing manipulation traces detectable by human eyes.  It
becomes very challenging to distinguish Deepfake images from real ones
with human eyes against new generations of Deepfake technologies.
Furthermore, adding different kinds of perturbation (e.g., blur, noise
and compression) can hurt the detection performance of Deepfake
detectors since manipulation traces are mixed with perturbations.  A
robust Deepfake detector should be able to tell the differernces between
real images and fake images generated by the GAN techniques although
both of them experience such modifications. 

There have been three generations of fake video datasets created for the
research and development purpose. They demonstrate the evolution of
Deepfake techniques. The UADFV dataset \citep{li2018exposing} belongs to
the first generate. It has only 50 video clips generated by one Deepfake
method. Its real and fake videos can be easily detected by humans.
FaceForensics++ \citep{rossler2019faceforensics++} and Celeb-DF-v2
\citep{li2020celeb} are examples of the second generation. They contain
more video clips with more identities. It is difficult for humans to
distinguish real and fake faces for them.  DFDC
\citep{dolhansky2020deepfake} is the third generation dataset. It
contains more than 100K fake videos which are generated by 8 Deepfake
techniques and perturbed by 19 kinds of distortions. The size of the
third generation dataset is very large. It is designed to test the
performance of various Deepfake detectors in an environment close to
real world applications. 

\begin{figure*}[!t]
\centering
\includegraphics[width=0.95\linewidth]{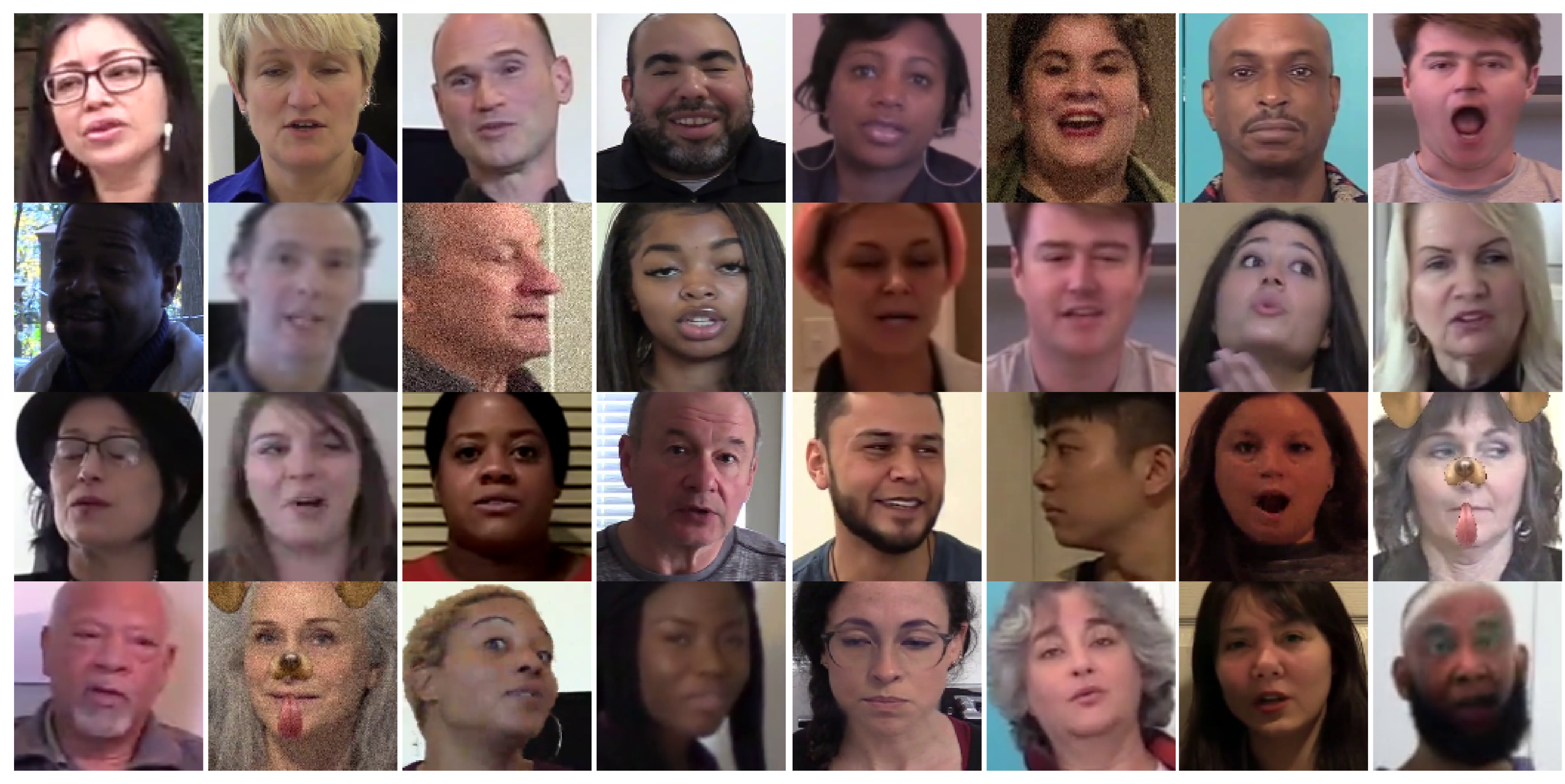}
\caption{Visualization of real and fake faces extracted from the third
generation DFDC dataset \citep{dolhansky2020deepfake}. The left and
right four columns depict real and fake faces, respectively.  Eight
Deepfake techniques are used to generate fake videos. Furthermore, 19
perturbations are added to real and fake videos.  Exemplary
perturbations include compression, additive noise, blur, change of
brightness, contrast and resolution, and overlay with flower and dog
patterns, random faces and images. A good Deepfake detector should be
able to distinguish real and fake videos with or without perturbation.}
\label{fig_exam}
\end{figure*}

State-of-the-art Deepfake detectors usually use the deep neural networks
(DNNs) to solve the Deepfake detection problem. Although they offer high
detection performance, their model sizes are so large that they cannot
be deployed on mobile phones. For example, the winning team of the DFDC
contest \citep{seferbekov2020dfdc} used seven pre-trained EfficientNets
that contain 432 million parameters.  In this research, our goal is to
develop a lightweight model with its size less than 256K. For a machine
learning model with size less than 256K, it can be run in any terminal
device with a limited amount of memory such as Raspberry Pi. 

A lightweight Deepfake detector called DefakeHop was developed in
\citep{chen2021defakehop}. An enhanced version of DefakeHop, called
DefakeHop++, is proposed in this work.  The improvements lie in two
areas.  First, DefakeHop examines three facial regions (i.e., two eyes
and mouth) only.  DefakeHop++ includes eight more landmark regions to
offer more information of human faces. Second, DefakeHop uses an
unsupervised energy criterion to select discriminant features.  It does
not exploit the correlation between features and labels.  DefakeHop++
adopts a supervised tool, called the Discriminant Feature Test (DFT), in
feature selection. The latter is more effective than the former due to
supervision. In DefakeHop++, spatial and spectral features from multiple
facial regions and landmarks are generated automatically and, then, DFT
is used to select a subset of discriminant features to train a
classifier. As compared with MobileNet v3, which a lightweight CNN model
of 1.5M parameters targeting at mobile applications, DefakeHop++ has an
even smaller model size of 238K parameters (i.e., 16\% of MibileNet v3).
In terms of Deepfake image detection performance, DefakeHop++
outperforms MobileNet v3 without data augmentation and leverage of
pre-trained models. 

The rest of the paper is organized as follows. Related work is reviewed
in Sec. \ref{sec:review}. DefakeHop++ is described in detail in Sec.
\ref{sec:method}. Experimental results are shown in Sec.
\ref{sec:experiments}.  Finally, concluding remarks are given in Sec.
\ref{sec:conclusion}. 

\section{Review of Related Work}\label{sec:review}

{\bf Deepfake Detection.} Most state-of-the-art Deepfake detection
methods use DNNs to extract features from faces. Their models are
trained with heavy augmentation (e.g., deleting part of faces) to
increase the performance.  Despite the high performance of these models,
their model sizes are usually very large. They have to be pre-trained by
other datasets in order to converge.  Several examples are given below.
The model of the winning team of the DFDC Kaggle challenges
\citep{seferbekov2020dfdc} has 432M parameters. Heo {\em et al.}
\citep{heo2021deepfake} improved this model by concatenating it with a
Vision Transformer(VIT), which has 86M parameters. Zhao {\em et al.}
\citep{zhao2021multi} proposed a model that exploits multiple spatial
attention heads to learn various local parts of a face and the textural
enhancement block to learn subtle facial artifacts.  To reduce the model
size and improve the efficiency, Sun {\em et al.}
\citep{sun2021improving} proposed a robust method based on the change of
landmark positions in a video. These landmarks are calibrated by the
neighbor frames. Afterwards, a two-stream RNN is trained to learn the
temporal information of the landmark position.  Since they did not
consider the image information and only consider the position
information, the model size is 0.18M which is relatively small.  Tran
{\em et al.} \citep{tran2021high} applied MobileNet to different facial
regions and InceptionV3 to the entire faces. Although this model is
smaller, it still demands 26M parameters. 

{\bf DefakeHop.} To address the challenge of the need of huge model
sizes and training data, a new machine learning paradigm called green
learning has been developed in the last six years
\citep{kuo2016understanding, kuo2019interpretable, chen2019pixelhop,
chen2020pixelhop++}. Its main goal is to reduce the model size and
training time while keep high performance. Green learning has been
applied to different applications, e.g., \citep{liu2021voxelhop,
monajatipoor2021berthop, zhang2021anomalyhop, rouhsedaghat2021low,
rouhsedaghat2021facehop, zhang2020unsupervised, kadam2021r,
zhang2020pointhop, zhang2020pointhop++}. Based on green learning,
DefakeHop was developed in \citep{chen2021defakehop} for the Deepfake
detection task.  It first extracts a large number of features from three
facial regions using PixelHop++ \citep{chen2019pixelhop} and then
refines them using feature distillation modules. Finally, it feeds
distilled features to the XGBboost classifier \citep{chen2016xgboost}.
The DefakeHop model has only 42.8K parameters, yet it outperforms many
DNN solutions in detection accuracy against the first and second
generataion Deepfake datasets. Recently, DefakeHop has been used
to detect fake satellite images in \citep{chen2021geo}.

\section{DefakeHop++}\label{sec:method}

\begin{figure*}[!t]
\centering
\includegraphics[width=1\linewidth]{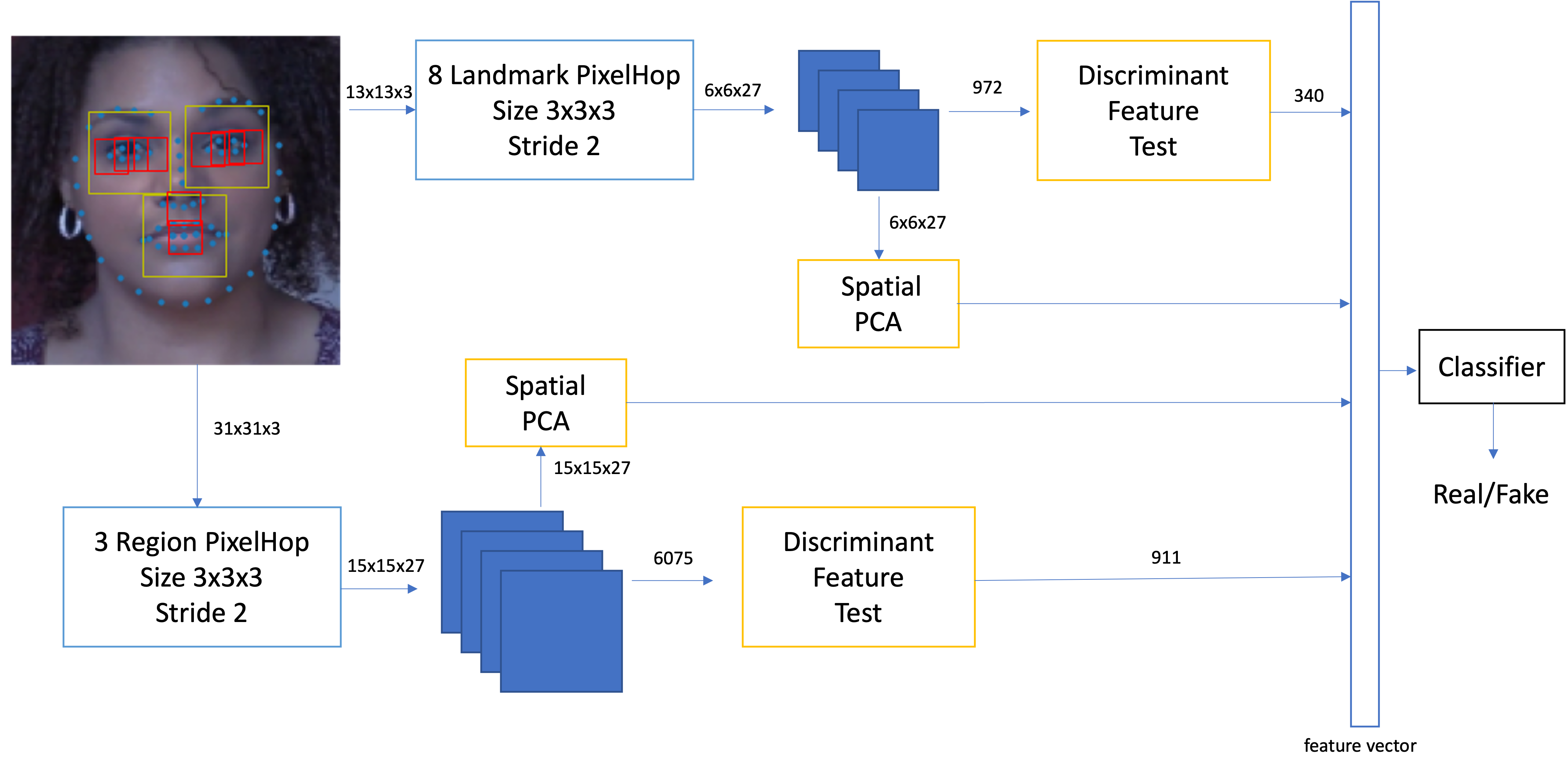}
\caption{An overview of DefakeHop++.}\label{fig_model}
\end{figure*}

DefakeHop++ is an improved version of DefakeHop. An overview of the
DefakeHop++ system is shown in Fig. \ref{fig_model}.  Facial blocks of
two sizes are first extracted from frames in a video sequence in the
pre-processing step. These blocks are then passed to DefakeHop++ for
processing. DefakeHop++ consists of four modules: 1) one-stage PixelHop,
2) spatial PCA, 3) discriminant feature test (DFT), and 4) Classifier.
The pre-processing step and the four modules are elaborated below. 

\subsection{Pre-processing Step}\label{subsec:preprocessing}

Frames are extracted from video sequences. For training videos, we
extract three frames per second. Since the video length is typically
around 10 seconds, we obtain around 30 frames per video. On the other
hand, the length and the frame number per second (FPS) are not fixed in
test videos. We uniformly sample 100 frames from each test video. 

\begin{figure}[!t]
\centering
\includegraphics[width=0.95\linewidth]{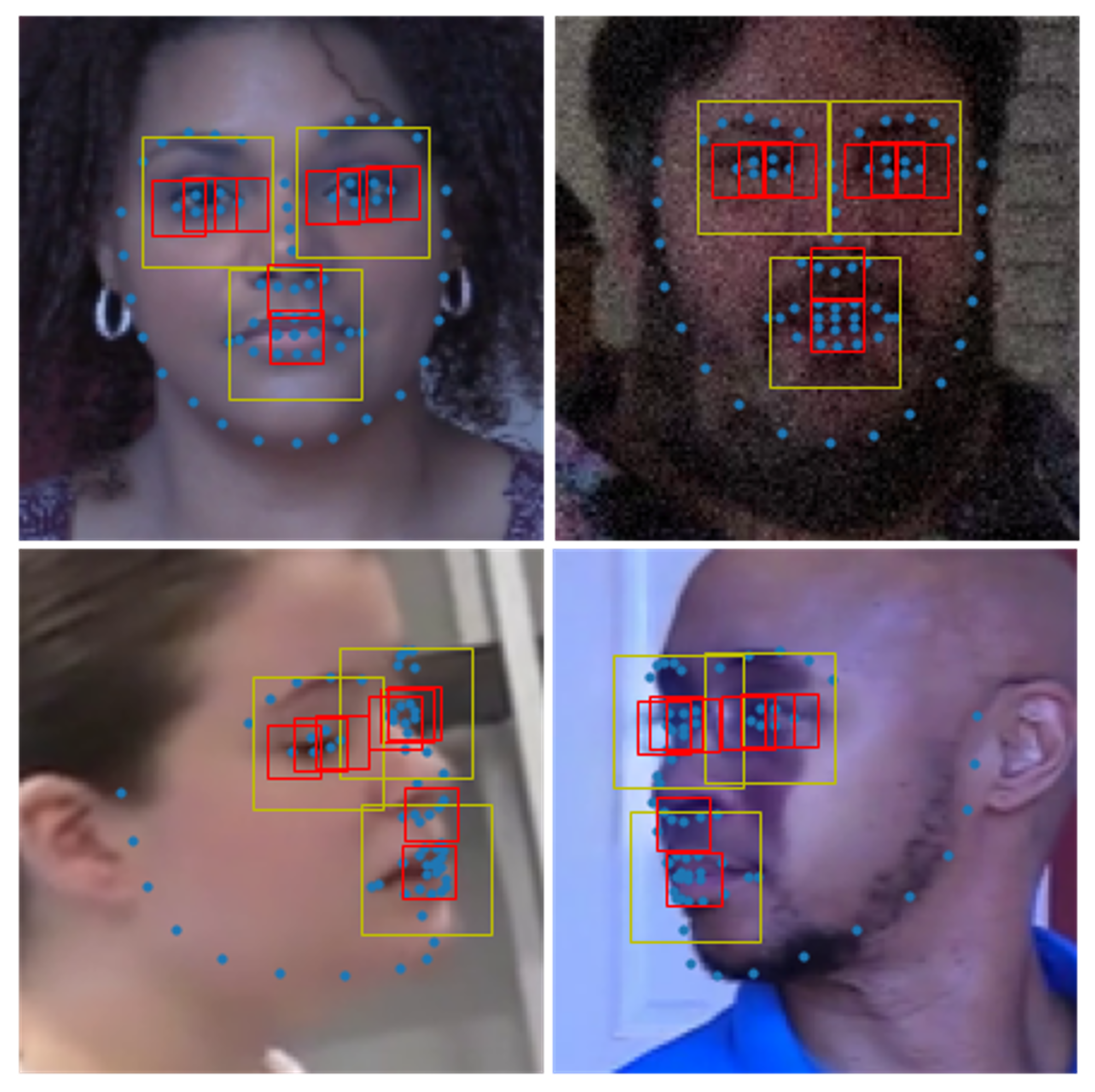}
\caption{Illustration of the prepossessing step. It first extracts 68 landmarks (in blue)
from the face. Then, it crops out blocks of size $31 \times 31$ from three regions
(in yellow) and blocks of size $13 \times 13$ from eight landmarks (in red). The step
can extract consistent blocks despite different head poses and perturbations. }\label{fig_pre}
\end{figure}

Facial landmarks are obtained by OpenFace2. With 68 facial landmarks, we
crop faces with 30\% of margin and resize them to $128 \times 128$ with
zero padding (if needed).  We do not align the face since face alignment
may distort the original face image and the side face is difficult to
align.  We conduct experiments on the discriminant power of each
landmark and find that two eyes and the mouth are most discriminant
regions.  Thus, we crop out 8 smaller blocks that cover 6 representative
landmarks from two eyes, one from the nose and one from the mouth.
Furthermore, we crop out three larger blocks to cover the left eye,
right eye and mouth.  We make the block size a hyper-parameter for user
to choose.  For the experiments reported in Sec. \ref{sec:experiments},
we adopt smaller blocks of size $13 \times 13$ centered at landmarks and
larger blocks of size $31 \times 31$ for three facial regions (i.e., two
eyes and the mouth) since they give the best results.  The extracted
small and large blocks are illustrated in Fig. \ref{fig_pre}. 

\subsection{Module 1: One-Stage PixelHop}

A PixelHop unit \citep{chen2019pixelhop, chen2020pixelhop++} contains a
set of filters used to extract features from training blocks. While the
filter weights of traditional filter banks (e.g., the Gabor or Laws
filters) are fixed, those of the PixelHop are data dependent. Its filter
banks are defined by the Saab transform, which decomposes a local
neighborhood (i.e., a patch) into one DC (direct current) and multiple
AC (alternated current) components. The DC component is the mean of each
patch. The AC components are obtained through the principal component
analysis (PCA) of patch residuals.  

To give an example, we set the patch size to $3 \times 3$ in the
experiment.  For the color face image input, each patch contains $(3
\times 3) \times 3$ degrees of freedom, including nine spatial pixel
locations and three spectral components. By collecting a large number of
patches, we can conduct PCA to obtain AC filters.  The eigenvectos and
the eigenvalues of the covariance matrix correspond to AC filters and
their mean energy values, respectively. Since most natural images are
smooth, the leading AC filters extract low-frequency features of higher
energy. When the frequency index goes higher, the energy of higher
frequency components bdecays quickly.  The Saab filter bank decouples the
27 correlated input dimensions into 27 decorrelated output dimensions.
Besides decorrelating input components, the Saab transform allows to
discard some high frequency channels due to their very small variances
(or energy values). 

The horizontal output size of a block can be calculated as
\begin{equation}
\frac{\mbox{horizontal block size} - \mbox{horizontal filter size}}{s} + 1,
\end{equation}
where $s$ is the stride parameter along the horizontal direction.  For
example, if the horizontal block size is 13, the horizontal filter size
is 3 and the horizontal stride is 2, then the horizontal output size is
equal to 6. The same computation applies to the vertical output size. 

\subsection{Module 2: Spatial PCA}\label{sec:spatial_PCA}

The output of the same frequency component may still have correlations
in the spatial domain.  This is especially true for the DC component and
low AC components. The correlation can be removed by spatial PCA. The
idea of spatial PCA is inspired by the eigenface method in
\citep{turk1991eigenfaces}. For each channel, we train a spatial PCA and
keep the leading components that have cumulative energy up to 80\% of
the total energy. This helps reduce the output feature dimension. 

Modules 1 and 2 describe the feature generation procedure for
DefakeHop++.  It is worthwhile to point out the differences in feature
generation of DefakeHop and DefakeHop++. 
\begin{itemize}
\item DefakeHop only focuses on two eyes and mouth three regions.
Besides these three regions, DefakeHop++ zooms into the neighborhood of
8 landmarks called a block to gain more detailed information. The justification of
these 8 landmarks is given in the last paragraph of Sec. \ref{subsec:setup}
\item DefakeHop conducts three-stage PixelHop units and applies spatial
PCA to the response output of all three stages. The pipeline is
simplfied to a one-stage PixelHop in DefakeHop++. Yet, the
simplification does not hurt the performance since the spatial PCA
applied to each block (or region) still offer the global information of
the corresponding block (or region). Note that such simplification is
needed as DefakeHop++ covers more spatial patches and regions. 
\end{itemize}

\subsection{Module 3: Discriminant Feature Test}

\begin{figure}[!t]
\centering
\includegraphics[width=\linewidth]{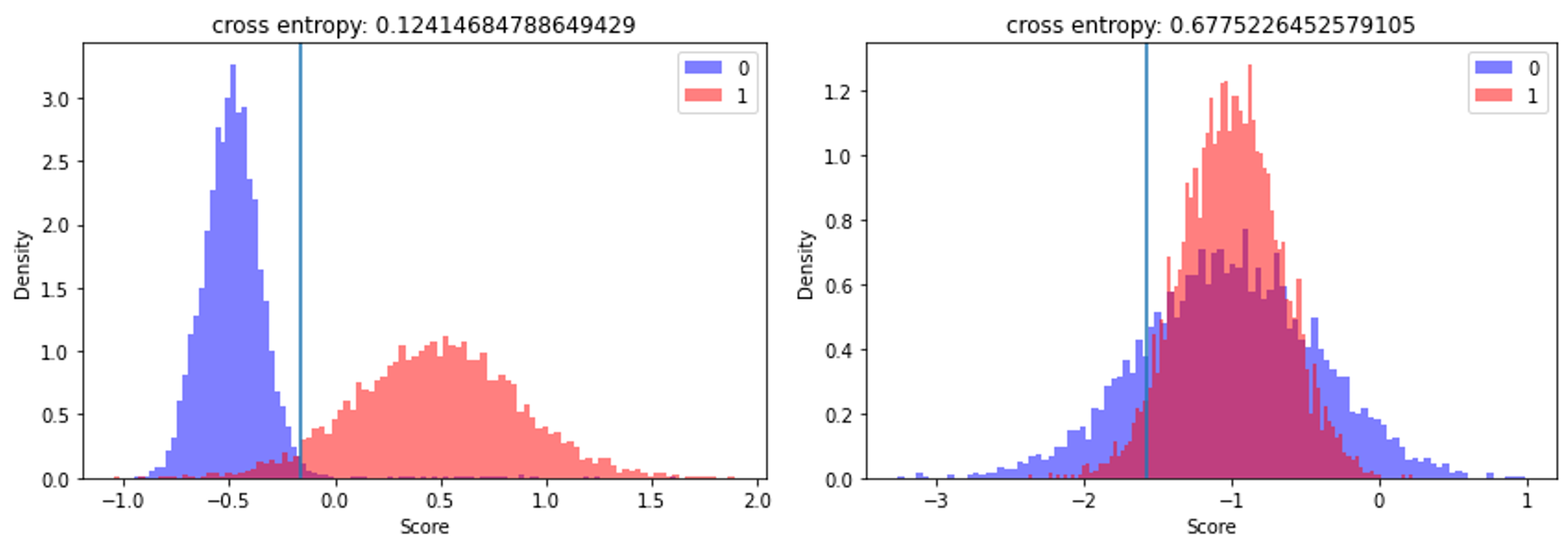}
\caption{Illustration of the feature selection idea in the discriminant
feature test (DFT). For the feature dimension associated with the left
subfigure, samples of class 0 and class 1 can be easily separated by the
blue partition line. Its cross entropy is lower. For the feature
dimension associated with the right subfigure, samples of class 0 and
class 1 overlap with each other significantly. It is more difficult to
separate them and its cross entropy is higher. Thus, the feature
dimension in the left subfigure is preferred.}\label{fig_dft}
\end{figure}

\begin{figure}[!t]
\centering
\includegraphics[width=\linewidth]{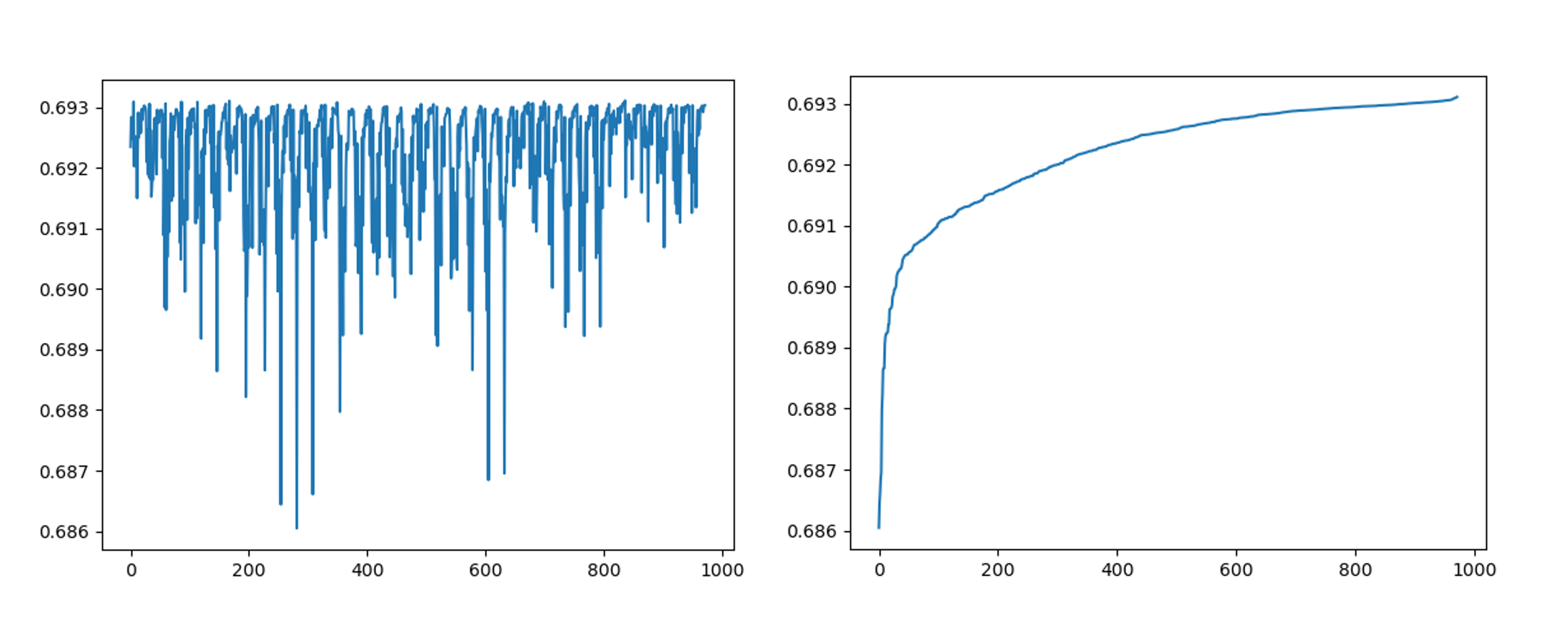}
\caption{Illustration of the feature selection process for a single
landmark, where the y-axis is the cross entropy of each dimension and
the x-axis is the channel index. The left and right subfigures show
unsorted and sorted feature dimensions.}\label{fig_selection}
\end{figure}

DefakeHop selects responses from channels of larger energy as features
into a classifier under the assumption that features of higher variance
are more discriminant. This is an unsupervised feature selection method.
Recently, a supervised feature selection method called the discriminant
feature test (DFT) was proposed in \citep{yang2022supervised}. DFT
provides a powerful method in selecting discriminant features using training
labels. The DFT process can be simply stated below. For each feature
dimension, we define an interval using its minimum and maximum values
across all training samples. Then, we partition the interval into two
sub-intervals at all possible split positions which are equally spaced
in the interval. For example, if we may select 31 split positions
uniformly distributed over the interval. For each partitioning, we can
use the maximum likelihood principle to assign a predicted label to each
training sample and compute the cross entropy of all training samples
accordingly. The split position that gives the lowest cross entropy is
chosen to be the optimal split of the feature and the associated
cross entropy is used as the cost function of this feature. The lower
the cross entropy, the most discriminant the feature dimension. We refer
to \citep{yang2022supervised} for more details.

We use Figs. \ref{fig_dft} and \ref{fig_selection} to explain the DFT
idea. Two features are compared in Fig. \ref{fig_dft}. The feature in
the left subfigure has a lower cross entropy value than the one in the
right subfigure. The left one is more discriminant than the right one,
which is intuitive.  Fig. \ref{fig_selection} is used to describe the
feature selection process for a given landmark block. The feature
dimension of one landmark block is 972. We perform DFT on each feature
and obtain 972 cross entropy values. The y-axis in both subfigures is
the cross entropy value while the x-axis is the channel index. In the
left subfigure, a smaller channel index indicates a lower frequency
channel. We see that discriminant channels of lower cross entropy
actually spread out in both low and high frequency channels. In the
right plot, we sort channels based on their cross entropy values, which
have an elbow point at 350. As a result, we can reduce the feature
number from 972 to 350 (35\%) by selecting 350 features with the lowest
cross entropy. 

\subsection{Module 4: Classification}

In DefakeHop, soft decisions from different regions are used to train
several XGBoost classifiers. Then, another ensemble XGBoost classifier
is trained to make the final decision. This is a two-stage decision
process. We find that a lot of detailed information is lost in the
first-stage soft decisions and, as a result, the ensemble classification
idea is not effective. To address this issue, we simply collect all
feature vectors from different regions and landmarks, apply DFT for
discriminant feature selection and train a LightGBM classifier in the
last stage in DefakeHop++. LightGBM and XGBoost play a similar role. The
main difference between DefakeHop and DefakeHop++ is that the former is
a two-stage decision process while the latter is a one-stage decision
process.  The one-stage decision process can consider the complicated
relations of features in landmarks and regions across all frequency
bands.  Once all frame-level predictions for a video are obtained, we
simply use their mean as the final prediction for the video. 

\begin{sidewaystable}[ht]
\caption{Comparison of detection performance of several methods on the
first genreation datasets with AUC as the performance metric.  The AUC
results of DefakeHop++ in both frame-level and video-level are given.
The best and the second-best results are shown in boldface and
underbared, respectively.  The AUC results of benchmarking methods are
taken from \cite{li2020celeb} and the number of parameters are from
\url{https://keras.io/api/applications}.  Also, we use $^{a}$ to denote
deep learning methods and $^{b}$ to denote non-deep-learning
methods.}\label{tab:compare1}
\begin{center}
\begin{tabular}{c|c|cc|c} \hline\hline
                                                &                                               & \multicolumn{2}{c|}{1st Generation}  &  \\ \hline
Method                                          & Model                                         & UADFV     & FF++           & \makecell{\#param} \\ \hline
Two-stream \citep{zhou2017two}                  & InceptionV3$^{a}$\citep{szegedy2016rethinking}& 85.1\%    & 70.1\%            & 23.9M \\
Meso4 \citep{afchar2018mesonet}                 & Designed CNN$^{a}$                            & 84.3\%    & 84.7\%            & 28.0K \\
MesoInception4 \citep{afchar2018mesonet}        & Designed CNN$^{a}$                            & 82.1\%    & 83.0\%            & 28.6K \\
HeadPose \citep{yang2019exposing}               & SVM$^{b}$                                     & 89.0\%    & 47.3\%            & - \\
FWA \citep{li2018exposing}                      & ResNet-50$^{a}$\citep{he2016deep}             & 97.4\%    & 80.1\%            & 25.6M \\
VA-MLP \citep{matern2019exploiting}             & Designed CNN$^{a}$                            & 70.2\%    & 66.4\%            & - \\
VA-LogReg \citep{matern2019exploiting}          & Logistic Regression$^{b}$                     & 54.0\%    & 78.0\%            & - \\
Xception-raw \citep{rossler2019faceforensics++} & XceptionNet$^{a}$\citep{chollet2017xception}  & 80.4\%    & \underbar{99.7\%}            & 22.9M \\
Xception-c23 \citep{rossler2019faceforensics++} & XceptionNet$^{a}$\citep{chollet2017xception}  & 91.2\%    & \underbar{99.7\%}            & 22.9M \\
Xception-c40 \citep{rossler2019faceforensics++} & XceptionNet$^{a}$\citep{chollet2017xception}  & 83.6\%    & 95.5\%            & 22.9M \\
Multi-task \citep{nguyen2019multi}              & Designed CNN$^{a}$                            & 65.8\%    & 76.3\%            & - \\
Capsule \citep{nguyen2019use}                   & CapsuleNet$^{a}$\citep{sabour2017dynamic}     & 61.3\%    & 96.6\%            & 3.9M \\
DSP-FWA \citep{li2019exposing}                  & SPPNet$^{a}$\citep{he2015spatial}             & \underbar{97.7\%}    & 93.0\%            & - \\
Multi-attentional \citep{zhao2021multi}         & Efficient-B4$^{a}$\citep{tan2019efficientnet} & -         & {\bf 99.8\%}            & 19.5M \\
DefakeHop \citep{chen2021defakehop}             & DefakeHop$^{b}$                               & {\bf 100\%}     & 96.0\%            & 42.8K \\
Ours (Frame Level)                              & DefakeHop++$^{b}$                             & {\bf 100\%}     & 98.4\%            & 238K \\
Ours (Video Level)                              & DefakeHop++$^{b}$                             & {\bf 100\%}     & 99.3\%            & 238K \\\hline

\end{tabular}
\end{center}
\end{sidewaystable}

\begin{sidewaystable}[ht]
\caption{Comparison of detection performance of several Deepfake
detectors on the second genreation datasets under cross-domain training
and with AUC as the performance metric.  The AUC results of DefakeHop
anad DefakeHop++ in both frame-level and video-level are given. The best
and the second-best results are shown in boldface and underbared,
respectively. Furthermore, we include results of DefakeHop and
DefakeHop++ under the same-domain training in the last 4 rows.  The AUC
results of benchmarking methods are taken from \cite{li2020celeb} and
the number of parameters are from
\url{https://keras.io/api/applications}.  Also, we use $^{a}$ to denote
deep learning methods and $^{b}$ to denote non-deep-learning
methods.}\label{tab:compare2}
\begin{center}
\begin{tabular}{c|c|cc|c} \hline\hline
                                                &                                               & \multicolumn{2}{c|}{2nd Generation}  &   \\ \hline
Method                                          & Model                                         & Celeb-DF v1     & Celeb-DF v2 & \makecell{\#param} \\ \hline
Two-stream \citep{zhou2017two}                  & InceptionV3$^{a}$& 55.7\%    & 53.8\%            & 23.9M \\
Meso4 \citep{afchar2018mesonet}                 & Designed CNN$^{a}$                            & 53.6\%    & 54.8\%            & 28.0K \\
MesoInception4 \citep{afchar2018mesonet}        & Designed CNN$^{a}$                            & 49.6\%    & 53.6\%            & 28.6K \\
HeadPose \citep{yang2019exposing}               & SVM$^{b}$                                     & 54.8\%    & 54.6\%            & - \\
FWA \citep{li2018exposing}                      & ResNet-50$^{a}$             & 53.8\%    & 56.9\%            & 25.6M \\
VA-MLP \citep{matern2019exploiting}             & Designed CNN$^{a}$                            & 48.8\%    & 55.0\%            & - \\
VA-LogReg \citep{matern2019exploiting}          & Logistic Regression$^{b}$                     & 46.9\%    & 55.1\%            & - \\
Xception-raw \citep{rossler2019faceforensics++} & XceptionNet$^{a}$  & 38.7\%    & 48.2\%            & 22.9M \\
Xception-c23 \citep{rossler2019faceforensics++} & XceptionNet$^{a}$  & -    & 65.3\%            & 22.9M \\
Xception-c40 \citep{rossler2019faceforensics++} & XceptionNet$^{a}$  & -    & 65.5\%            & 22.9M \\
Multi-task \citep{nguyen2019multi}              & Designed CNN$^{a}$                            & 36.5\%    & 54.3\%            & - \\
Capsule \citep{nguyen2019use}                   & CapsuleNet$^{a}$     & -    & 57.5\%            & 3.9M \\
DSP-FWA \citep{li2019exposing}                  & SPPNet$^{a}$            & -    & \underbar{64.6\%}      & - \\
Multi-attentional \citep{zhao2021multi}         & Efficient-B4$^{a}$      & -    & {\bf 67.4\%}           & 19.5M \\

Ours (Frame Level)              & DefakeHop++$^{b}$    & \underbar{56.30\%}     & 60.5\%            & 238K \\
Ours (Video Level)              & DefakeHop++$^{b}$    & {\bf 58.15\%}     & 62.4\%            & 238K \\\hline

Ours (Trained on Celeb-DF, Frame Level)             & DefakeHop$^{b}$      & 93.1\%    & 87.7\% & 42.8K \\
Ours (Trained on Celeb-DF, Video Level)             & DefakeHop$^{b}$      & 95.0\%    & 90.6\% & 42.8K \\
Ours (Trained on Celeb-DF, Frame Level)             & DefakeHop++$^{b}$    & \underbar{95.4\%} & \underbar{94.3\%}  & 238K \\
Ours (Trained on Celeb-DF, Video Level)             & DefakeHop++$^{b}$    & {\bf 97.5\%}    & {\bf 96.7\%}         & 238K \\\hline
\end{tabular}
\end{center}
\end{sidewaystable}

\section{Experiments}\label{sec:experiments}

We compare the detection performance of DefakeHop++, state-of-the-art
deep learning and non-deep learning methods on several datasets as well
as their model sizes and training time in this section to demonstrate
the effectiveness of DefakeHop++.

\subsection{Experimental Setup}\label{subsec:setup}

{\bf Datasets.} Deepfake video datasets can be categorized into three
generations based on the dataset size and Deepfake methods used for fake
image generation. The first-generation datasets includes UADFV and FF++.
The second-generation datasets include Celeb-DF version 1 and version 2.
The third-generation dataset is the DFDC dataset.  The datasets of later
generations have more identities of different races, utilize more
deepfake algorithms to generate fake videos, and add more perturbation
types to test videos. Apparently, the later generation is more
challenging than the earlier generation. The datasets used in our
experiments are described below.

\begin{itemize}
\item {\bf UADFV} \citep{li2018exposing} \\
UADFV is the first Deepfake detection dataset. It consists of 49 real
videos and 49 fake videos.  Real videos are collected from YouTube while
fake ones are generated by the FakeApp mobile App \citep{Fakeapp}.
\item {\bf FaceForensics++ (FF++)} \citep{rossler2019faceforensics++} \\
It contains 1000 real videos collected from YouTube. Two popular
methods, FaceSwap \citep{Kowalski2016} and Deepfakes
\citep{Deepfakes2018}, are used in fake video generation. Each of them
generated 1000 fake videos of different quality levels, e.g., RAW, HQ
(high quality) and LQ (low quality). In the experiment, we focus on HQ
compressed videos since they are more challenging to detect by Deepfake
detection algorithms. 
\item {\bf Celeb-DF} \citep{li2020celeb} \\
Celeb-DF has two versions. Celeb-DF v1 contains 408 real videos from
YouTube and 795 fake videos.  Celeb-DF v2 consists of 890 real and 5639
fake videos.  Fake videos are created by an advanced version of
DeepFake. These videos contain subjects of different ages, ethnicities
and sex. Celeb-DF v2 is a superset of Celeb-DF v1.  Celeb-DF v1 and v2
have been widely tested on many Deepfake detection methods. 
\item {\bf DFDC} \citep{dolhansky2020deepfake} \\
DFDC is the third generation dataset. It contains more than 100K videos
generated by 8 different Deepfake algorithms. The test videos are
perturbed by 19 distractors and augmenters such as change of
brightness/contrast, logo overlay, dog filter, dots overlay, faces
overlay, flower crown filter, grayscale, horizontal flip, noise, images
overlay, shapes overlay, change of coding quality level, rotation, text
overlay, etc. The dataset was generated to mimic the real-world application 
scenario. It is the most challenging one among the four benchmark datasets. 
\end{itemize}

{\bf Evaluation Metrics.} Each Deepfake detector assigns a probability
score of being a fake one to all test images. Then, these scores can be
used to plot the Receiver Operating Characteristic (ROC) curve.  Then,
the area under curve (AUC) score can be used to compare the performance
of different detectors. We report the AUC scores at the frame level as
well as the video level. 

\begin{figure}[!t]
\centering
\includegraphics[width=\linewidth]{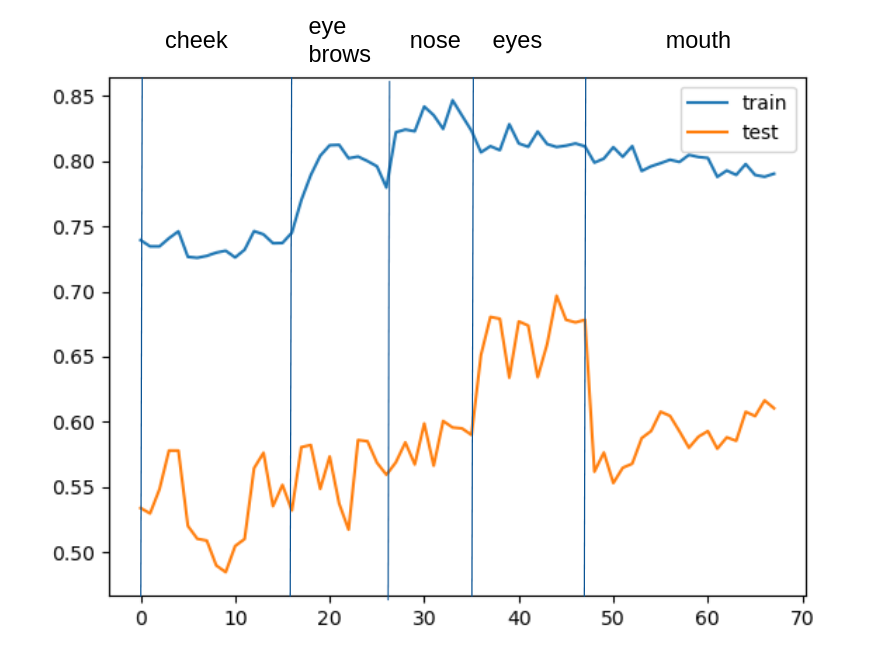}
\caption{Analysis of landmark discriminability, where the x-axis is the
landmark index anad the y-axis is the AUC score. Landmarks in the two
eye regions are most discriminant in both training and test datasets.}
\label{fig_reg}
\end{figure}

{\bf Discriminability Analysis of Landmarks.} As mentioned in Sec.
\ref{subsec:preprocessing}, there are 68 landmarks. We use the AUC
scores to analyze the discriminability of the 68 landmark regions in
Fig.  \ref{fig_reg}. We see from the figure that the performance of
different landmarks varies a lot. Landmarks in two eye regions are most
discriminant. This is attributed to the fact that eyes have rich details
and complex movement and, as a result, they cannot be well synthesized
by any Deepfake algorithms. The mouth region is the next discriminant
one because it is difficult to synthesize lip motion and teeth. The
cheek and nose regions are less discriminant since they are relatively
smooth in space and stable in time.  This justifies our choice of six
landmarks from eyes, one landmark from the nose and one landmark from
the mouth as described in Sec.  \ref{subsec:preprocessing}. 

\subsection{Detection Performance Comparison}

{\bf First Generation Datasets.} The performance of a few Deepfake
detectors on two first generation datasets, UADFV and FF++, is compared
in Table \ref{tab:compare1}. Both DefakeHop and DefakeHop++ achieve a
perfect AUC score of 100\% against UADFV. DSP-FWA and FWA are two closer
ones with AUC scores of 97.7\% and 97.4\%, respectively. Actually, UADFV
is an easy dataset where the visual artifacts are visible to human eyes.
It is interesting to see that the extremely large models do not reach
perfect detection results. This could be explained by the small size of
UADFV (i.e., 49 real videos and 49 fake videos). The model sizes of
DefakeHop and DefakeHop++ can be sufficiently trained by a small dataset
size. For FF++, Multi-attentional achieves the best AUC score (i.e., 99.8\%)
while Xception-raw and Xception-c23 achieves the second best (i.e.,
99.7\%). DefakeHop++ with its AUC computation at the videl level has the
next highest AUC score (i.e., 99.3\%).  The performance gap among them
is small. Furthermore, the performance of Xception-raw and Xception-c23
is boosted by a larger dataset size of FF++, which has 1000 real videos
of different quality levels. 

{\bf Second Generation Datasets with Cross-Domain Training.} The
performance of several Deepfake detectors, which are trained on the FF++
dataset, against two second generation datasets is compared in Table
\ref{tab:compare2}. We see that video-level DefakeHop++ gives the best
AUC score while frame-level DefakeHop++ gives the second best AUC score
for Celeb-DF-v1. As to Celeb-DF-v2, Multi-attentional yield the best AUC
score while Xception-raw and Xception-c23 offer the second best scores.
DefakeHop++ is slightly inferior to them. Furthermore, we show the
performance of DefakeHop and DefakeHop++ under the same domain training
in the last four rows of Table \ref{tab:compare2}. Their performance
has improved significantly. Video-level DefakeHop++ outperforms video-level 
DefakeHop by 2.5\% in Celeb-DF v1 and 6.1\% in Celeb-DF v2. 

\begin{figure}[!t]
\centering
\includegraphics[width=0.9\linewidth]{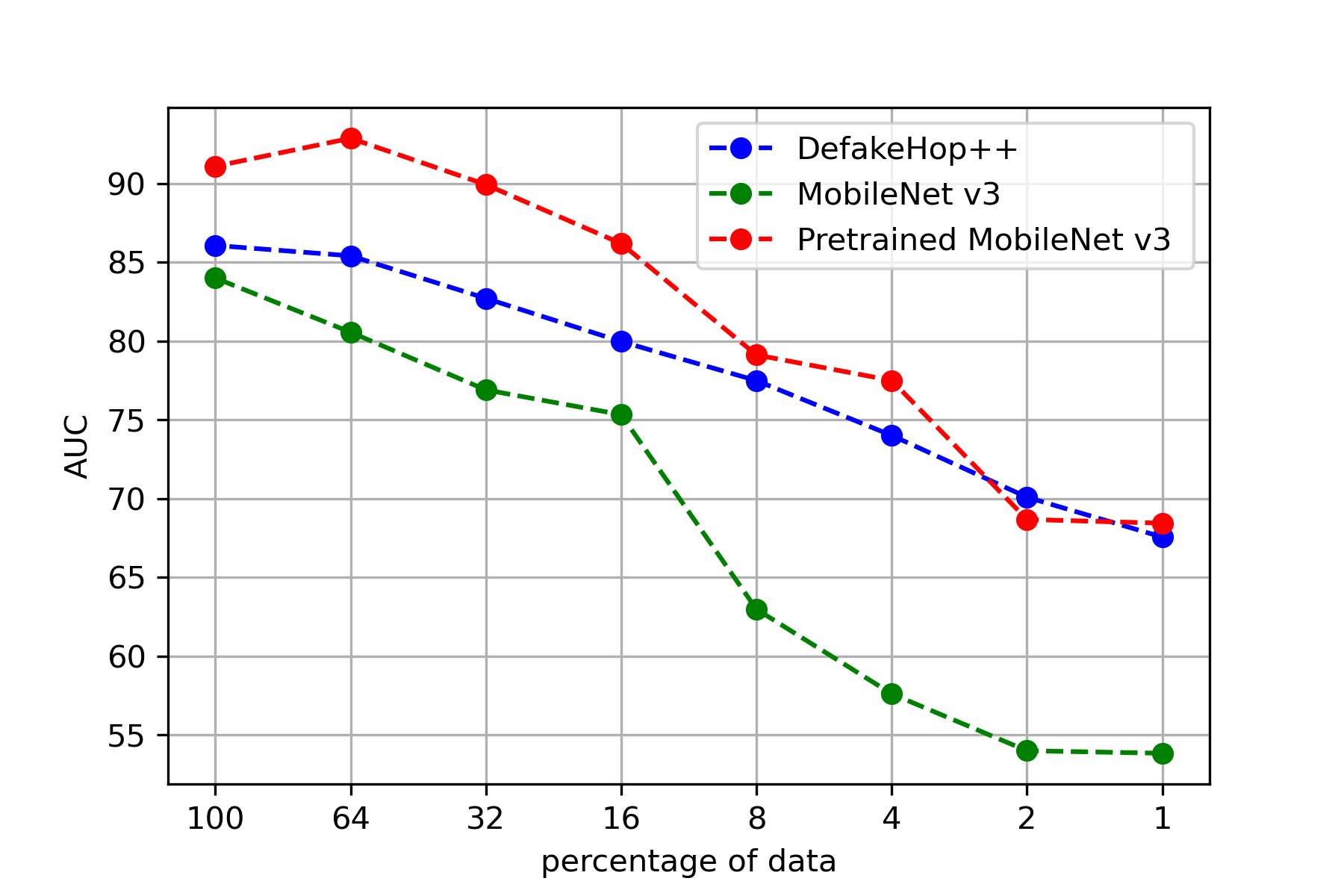}
\caption{Detection performance comparison of DefakeHop++, MobileNet v3
with pre-training by ImageNet and MobileNet v3 without pre-training as a
function of training data percentages of the DFDC dataset.}\label{fig_weak}
\end{figure}

\begin{figure}[!b]
\centering
\includegraphics[width=0.9\linewidth]{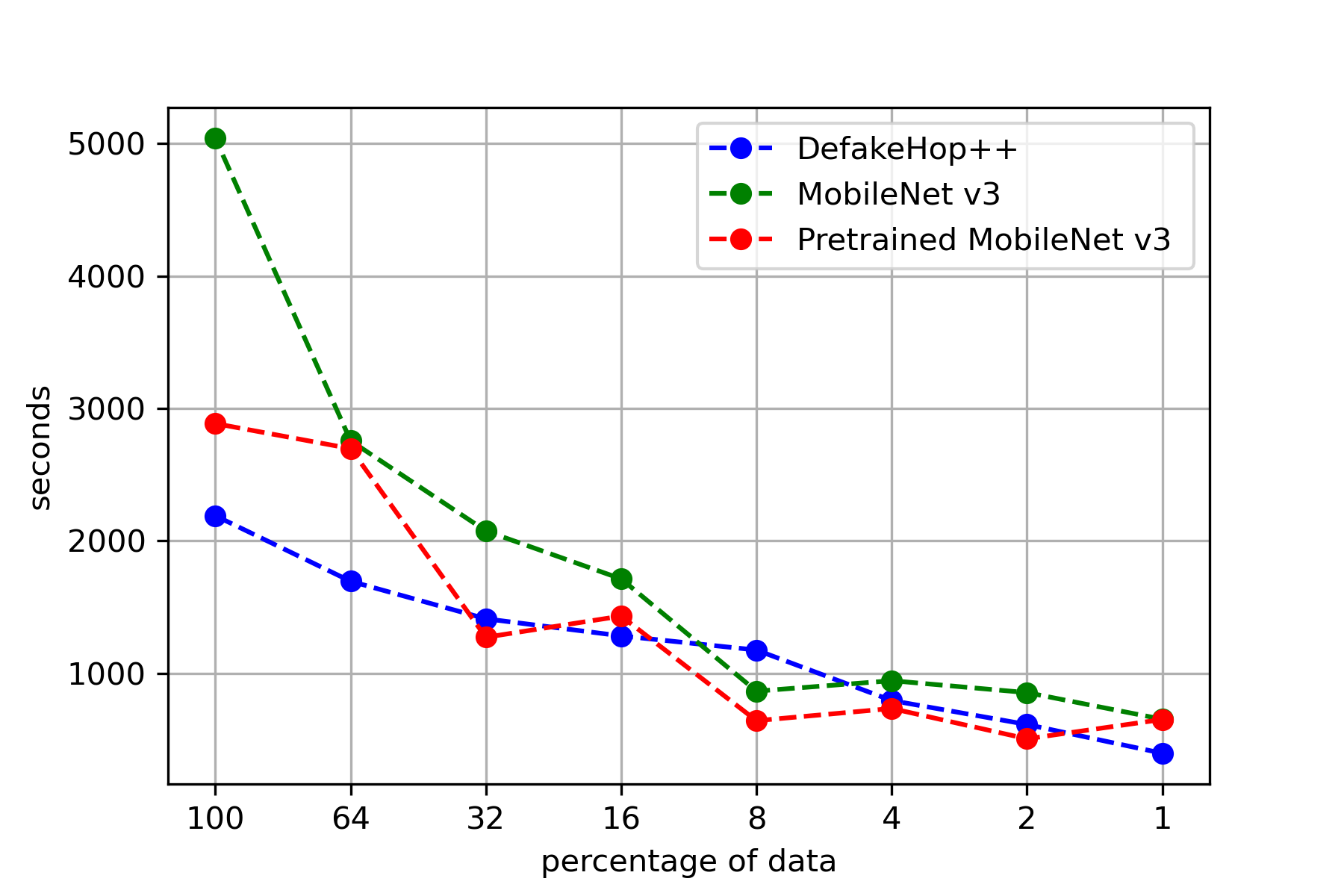}
\caption{Training time comparison of DefakeHop++, MobileNet v3 with
pre-training by ImageNet and MobileNet v3 without pre-training as a
function of training data percentages of the DFDC dataset, where the
training time is in the unit of seconds. The training time does not
include that used in the pre-processing step.}\label{fig_time}
\end{figure}

{\bf Third Generation Dataset.} The size of the third generation
dataset, DFDC, is huge. It demands a lot of computational resources
(including training hardware, time and large models) to achieve high
performance.  Since our main interest is on lightweight detection
algorithms, we focus on the comparison of DefakeHop++ and MobileNet v3,
which has 1.5M parameters and targets at mobile applications.  We train
both models with parts and/or all of DFDC training data and report the
detection performance on the test dataset in Fig.  \ref{fig_weak}.  We
have the following three observations from the figure. First,
pre-trained MobileNet v3 gives the best result, DefakeHop++ the second,
and MobileNet v3 without pre-training the worst. It shows that, if there
are sufficient training data in training, the detection performance of a
larger model can be boosted. For the same reason, the detection
performance of all three models decreases as the training data of DFDC
becomes less. Second, with 1-8\% of DFDC training data, the performance
of DefakeHop++ and pre-trained MobileNet v3 is actually very close. The
performance gap between DefakeHop++ and MobileNet v3 without
pre-training is significant in all training data ranges. For example,
with only 1\% of the DFDC training data, the AUC score of DefakeHop++
reaches 68\% while that of MobileNet V3 can only reach 54\%. Third, with
100\% of the DFDC training data but without any data augmentation,
DefakeHop++ still can achieve an AUC score of 86\%, which is 5\% lower
than pre-trained MobileNet v3. 

Furthermore, we compare the training time on the three models as a
function of the training data percentage in Fig.  \ref{fig_time}.  The
model is trained on AMD Ryzen 9 5950X with Nvidia GPU 3090
24G.  If a CNN is not pre-trained, it generally needs more time to
converge.  The training time of DefakeHop++ is lowest for 64\% and 100\%
of the total training data. Its training time is about the same as the
pre-trained MobileNet v3 for the cases of 1-32\% training data. 

\subsection{Model Size of DefakeHop++} 

DefakeHop++ consists of one-stage PixelHop, Spatial PCA, DFT and
Classifier modules. The size of each component can be computed as
follows. 

{\bf PixelHop.} The parameters are the filter weights. Each filter
has a size of $(3 \times 3) \times 3 =27$. Since there are 27 filters, the
total number of parameters of PixelHop is $27\times 27=729$ parmaeters. 

{\bf Spatial PCA.} For each channel, we flatten 2D spatial responses to
a 1D vector and train a PCA. We conduct spatial PCA on channels of
higher energy (i.e. those with cumulative energy up to 80\% total
energy).  Furthermore, we set an upper limit of 10 channels to avoid a
large number of filters for a particular channel. Based on this design
guideline, the averaged channel numbers for a landmark and a spatial
region are 35 and 40, respectively. 

{\bf DFT.} DFT is used to select a subset of discriminant features. Its
parameters are indices of selected channels. For 8 landmark blocks, we
keep features in top 35\%. For 3 spatial regions, we keep features in
top 15\%. Then, the number of parameters of DFT is $(8 \times 340)+ (3
\times 911) = 5,453$ as shown in Table \ref{tab:parameters}. 

{\bf Classifier.} We use LightGBM \citep{ke2017lightgbm} as the
classifier and set the maximum number of leaves to 64. As a result, the
maximum intermediate node number of a tree is bounded by 63.  We store
two parameters (i.e., the selected dimension and its threshold) at each
intermediate node and one parameter (i.e., the predicted soft decision
score) at each leaf node. The number of parameters for one tree is
bounded by 190. Furthermore, we set the maximum number of tree to 1000.
Thus, the number of parameters for LightGBM is bounded by 190K. 

\begin{table}[ht]
\vspace{-0.3cm}
\caption{The number of parameters for various parts.}\label{tab:parameters}
\begin{center}
\vspace{0.1cm}
\begin{tabular}{c|c|c|c|c} \hline \hline
  & Subsystem & Number & Parameters & Total\\ \hline
\multirow{2}{*}{Pixelhop} & Landmarks   & 8 & 3x3x3=729 & 5832\\ 
                          & Regions     & 3 & 3x3x3=729 & 2187\\ \hline
\multirow{2}{*}{Spatial PCA} & Landmarks  & 8 & 6x6x35=1,260 & 10,080 \\ 
                             & Regions    & 3 & 15x15x40=9,000 & 27,000 \\ \hline
\multirow{2}{*}{DFT}      & Landmarks   & 8 & 6x6x27x0.35=340 & 2720\\ 
                          & Regions     & 3 & 15x15x27x0.15=911 & 2733\\ \hline
\multirow{1}{*}{LightGBM} & - & 1 & 190,000 & 190,000\\ \hline
\textbf{Total} & & & &\textbf{237,832}\\ \hline
\end{tabular}
\end{center}
\vspace{-0.2cm}
\end{table}

\section{Conclusion and Future Work}\label{sec:conclusion}

A lightweight Deepfake detection method, called DefakeHop++, was
proposed in this work. It is an enhanced version of our previous
solution called DefakeHop. Its model size is significantly smaller than
that of state-of-the-art DNN-based solutions, including MobileNet v3,
while keeping reasonably high detection performance.  It is most
suitable for Deepfake detection in mobile/edge devices. 

Fake image/video detection is an important topic. The faked content is
not restricted to talking head videos. There are many other application
scenarios. Examples include faked satellite images, image splicing,
image forgery in publications, etc. Heavyweight fake image detection
solutions are not practical. Furthermore, fakes images can appear in
many forms.  On one hand, it is unrealistic to include all possible
perturbations in the training dataset under the setting of heavy
supervision. On the other hand, the performance could be quite poor with
little supervision.  It is essential to find a midground and look for a
lightweight weakly-supervised solution with reasonable performance. This
paper shows our research effort along this direction. We will continue
to explore and generalize the mothodology to other challenging Deepfake
problems. 

\printbibliography

\end{document}